\documentclass[runningheads]{llncs}
\usepackage[T1]{fontenc}
\usepackage{graphicx}
\usepackage{tikz}
\usepackage{graphicx}
\usepackage{subcaption} 
\usepackage{amsmath}
\usepackage{amssymb}
\usepackage{hyperref}
\usepackage{comment}

\usetikzlibrary{shapes, arrows.meta, positioning, fit, calc, shadows}
\begin{document}
\title{FoMo‑X: Modular Explainability Signals for Outlier Detection Foundation Models}
\titlerunning{FoMo‑X: Modular Explainability for Outlier Foundation Models}
%
\author{Simon Klüttermann\inst{1}\orcidID{0000-0001-9698-4339}\thanks{Equal contribution.} \and
Tim Katzke\inst{1,2}\orcidID{0009-0000-0154-7735}\protect\footnotemark[1] \and
Phuong Huong Nguyen\inst{1}\orcidID{0009-0009-2144-144X} \and
Emmanuel Müller\inst{1,2}\orcidID{0000-0002-5409-6875}}
\authorrunning{S. Klüttermann et al.}

\institute{TU Dortmund University, Dortmund, Germany \and
Research Center Trustworthy Data Science and Security, University Alliance Ruhr, Dortmund, Germany\\
\email{Simon.Kluettermann@cs.tu-dortmund.de (corresponding author)}\\
\email{tim.katzke@tu-dortmund.de}
}
\maketitle  
\begin{abstract}
Tabular foundation models, specifically Prior-Data Fitted Networks (PFNs), have revolutionized outlier detection (OD) by enabling unsupervised zero-shot adaptation to new datasets without training. However, despite their predictive power, these models typically function as opaque black boxes, outputting scalar outlier scores that lack the operational context required for safety-critical decision-making. Existing post-hoc explanation methods are often computationally prohibitive for real-time deployment or fail to capture the epistemic uncertainty inherent in zero-shot inference. In this work, we introduce FoMo-X, a modular framework that equips OD foundation models with intrinsic, lightweight diagnostic capabilities.
We leverage the insight that the frozen embeddings of a pretrained PFN backbone already encode rich, context-conditioned relational information. FoMo-X attaches auxiliary diagnostic heads to these embeddings, trained offline using the same generative simulator prior as the backbone. This allows us to distill computationally expensive properties, such as Monte Carlo dropout based epistemic uncertainty, into a deterministic, single-pass inference. We instantiate FoMo-X with two novel heads: a Severity Head that discretizes deviations into interpretable risk tiers, and an Uncertainty Head that provides calibrated confidence measures. Extensive evaluation on synthetic and real-world benchmarks (ADBench) demonstrates that FoMo-X recovers ground-truth diagnostic signals with high fidelity and negligible inference overhead. By bridging the gap between foundation model performance and operational explainability, FoMo-X offers a scalable path toward trustworthy, zero-shot outlier detection.

\end{abstract}

\section{Introduction}
\label{sec:introduction}

Outlier detection (OD) is a core problem in machine learning, with diverse applications in safety-critical areas ranging from industrial monitoring~\cite{machinefault,faultapl2} to cybersecurity~\cite{auto_appl_httptraffic,auto_appl_networkintrusion} and healthcare~\cite{auto_appl_virusmutations,medapl}. In such deployments, practitioners rarely act on an outlier score alone: effective OD systems must also provide operationally meaningful signals that support downstream decisions such as escalation to human review, prioritization, or follow-up investigation~\cite{XADSurvey2appl,XADSurvey1general}.

This need has motivated a broad literature on explainable outlier detection~\cite{XADSurvey2appl,XADSurvey1general,metasurvey}, where common approaches include feature-centric attributions \cite{deanshap}, example- or prototype-based rationales~\cite{prototypeAD}, rule-based explanations, and counterfactual explanations~\cite{counterfactual4AD} that suggest minimal changes to render an instance nominal.
In parallel, classical OD methods such as an Isolation Forest~\cite{ifor} have been augmented with model-specific explainers~\cite{explainableIFOR}.

Recently, tabular foundation models based on prior-data fitted networks (PFNs) have introduced a new paradigm for tabular learning: instead of fitting a separate model per dataset, a transformer is pretrained under a simulator-induced task prior and adapts \emph{in-context} at inference time~\cite{tabpfn,tabpfnNature,Freiburg1}. FoMo-0D~\cite{fomo} applies this idea to context-conditioned outlier scoring, enabling zero-shot detection without per-dataset optimization; subsequent work further develops this line of OD foundation models~\cite{fomo2Outformer}. 
Although these models reduce common OD burdens such as hyperparameter tuning and model selection, their interpretability remains an emerging research frontier, as they typically expose only a hard-to-interpret scalar outlier score at test time, leaving users without intrinsic diagnostics for trust calibration and triage.

Computing such diagnostically valuable quantities post hoc can be costly. For instance, epistemic uncertainty via Monte Carlo dropout requires multiple stochastic forward passes, which is often impractical as an always-on layer~\cite{MCDropoutEpistemic}. In contrast, we argue that PFN internal representations provide the basis for more direct and computationally efficient intrinsic diagnostics and explanations.

Consequently, we propose FoMo-X (\textit{\underline{F}ear \underline{O}f \underline{M}issing \underline{O}utlier e\underline{X}planations}), a modular framework that augments PFN-style OD foundation models with lightweight, single-pass \emph{diagnostic heads}. Building on the observation that PFN-based detectors such as FoMo-0D compute a rich, context-dependent embedding for each query point, FoMo-X freezes the pretrained backbone and trains auxiliary readouts that decode operationally meaningful signals from this embedding, thereby preserving the original outlier predictions while adding almost no inference overhead. Head training leverages the same simulator prior used for PFN pretraining, enabling supervision not only for outlier labels but also for additional diagnostic targets that would be costly to compute online.

In this paper, we instantiate FoMo-X with two diagnostics: a \emph{severity} head that discretizes how strongly a sample deviates from context-defined normality, motivated by concerns that outlier scores are not necessarily severity-aligned \cite{positionpaperRoechnerBenchmarking} and by emerging multilevel OD benchmarks \cite{cao2024madbench}, and an \emph{uncertainty} head, that distills an epistemic-uncertainty proxy otherwise obtained via repeated stochastic forward passes (MC dropout) into a single deterministic prediction \cite{MCDropoutEpistemic}. While these outputs are not intended as causal feature attributions or counterfactual recourse\footnote{Because of the way OD foundation models are currently trained, it is almost impossible to access feature information. See Section~\ref{sec:discuss}.}, they provide actionable explainability signals that support trust calibration and escalation decisions in safety-critical OD deployments, complementing feature-level explanation methods when those are available and appropriate~\cite{katzke2025skinsplain,XADSurvey1general}. Conceptually, FoMo-X transfers established ideas from knowledge/dropout distillation~\cite{bulo2016dropoutdistill,gurau2018dropoutdistillation,hinton2015distillation} and representation probing on frozen backbones~\cite{alain2017probes} to the context-conditioned OD foundation-model setting.

Our main contributions are: (1) we propose a practical way to train outlier detection foundation models to explain themselves by introducing a head-based, modular approach, incorporating intrinsic diagnostic explainability signals without modifying the underlying detector, and (2) we instantiate this approach on top of FoMo-0D with an offline training protocol that relies solely on the model's context-conditioned embedding and simulator-derived supervision under negligible additional inference cost. (3) We develop two diagnostic heads for severity and uncertainty prediction, and (4) we evaluate their utility on synthetic tasks and real-world tabular OD benchmarks to show the potential and limitations of future modular foundation models for OD.

However, this represents only an initial step towards multi-headed explainability for outlier-detection foundation models; thus, we publish our trained model and the code to train additional heads at \url{https://github.com/psorus/FoMo-X}.

\section{Related Work}

\subsection{Prior-Data Fitted Networks and Tabular Foundation Models}

Prior-Data Fitted Networks (PFNs)~\cite{Freiburg1} are a class of transformers designed to perform In-Context Learning (ICL) on tabular data. Originally popularized by TabPFN~\cite{tabpfn} for classification, these models have achieved state-of-the-art (SOTA) performance across diverse domains~\cite{tabpfnNature,tabpfnRefine} and remain a focal point of theoretical and empirical research~\cite{tabfnRW,tabPFN2dot5,tabpfnTheory}.

Architecturally, PFNs are trained on a vast corpus of synthetic datasets to approximate the Posterior Predictive Distribution (PPD). Unlike traditional models that learn a solution for a specific dataset, PFNs learn the \textit{algorithm} for solving tasks within a model class. This paradigm has been successfully extended to regression~\cite{tabPFN2dot5} and, more recently, to outlier detection~\cite{fomo2Outformer,fomo}. A significant advantage of this approach in the OD context is its reliance on purely simulated data during the pre-training phase, which bypasses the scarcity of labeled real-world outliers.

Explainability for PFNs is an emergent research frontier~\cite{somewhatexplainTabPFN,hutterTabpfnIML} and, to our knowledge, remains largely unexplored for OD foundation models. While recent work begins to characterize what in-context tabular transformers implement and how their representations relate to classical, more interpretable model classes, the scale and architectural complexity of PFNs make faithful, low-overhead explanations non-trivial. Traditional post-hoc methods can be computationally expensive and technically challenging, and they often fail to match the simplicity of intrinsically interpretable tabular models~\cite{somewhatexplainTabPFN,hutterTabpfnIML}. We instead argue that PFNs can be extended beyond label prediction by incorporating diagnostic attributes into the simulator-based training objective, enabling the model to predict its own explanations via supervision from ground-truth logic defined with help of the synthetic data generator.

\subsection{Explainable Outlier Detection}

Outlier Detection (OD) is a mature field of machine learning with diverse applications ranging from fraud detection~\cite{auto_appl_electionfraud} and finance~\cite{auto_appl_finance} to fundamental science~\cite{mikuni}. The primary advantage of OD lies in its ability to operate without expensive labels for rare outlier classes~\cite{surveyzhao,metasurvey,survey-ruff}. Traditional methods~\cite{dean,ifor,dte} typically follow a two-step paradigm: extracting patterns from normal training data and subsequently validating these patterns against test samples. Recently, however, foundation models for OD~\cite{fomo2Outformer,fomo}, inspired by TabPFN~\cite{tabpfn}, have emerged. These models process training and test data simultaneously to predict outlierness in a single forward pass, achieving state-of-the-art (SOTA) performance with high operational efficiency.

Because OD often serves safety-critical domains where data collection is difficult or costly~\cite{XADSurvey2appl}, explainability is essential and has been surveyed extensively~\cite{XADSurvey1general,XADSurvey3images}.
A consistent theme, especially important for our analysis, is that OD explanation quality is difficult to evaluate in the unsupervised setting because explanatory ``ground truth'' is rarely defined, and deployed OD systems often require explanations that support triage and operational decision-making in safety-critical applications~\cite{XADSurvey2appl,XADSurvey1general}.

One direction therefore aims at \emph{intrinsic} or \emph{self-interpretable} detectors, either by constraining the detector class or by modifying existing algorithms to expose interpretable structure.
For instance, work on making the Isolation Forest algorithm~\cite{ifor} more interpretable~\cite{explainableIFOR} illustrates how detector-specific mechanisms can be turned into feature-relevant rationales without treating the detector as a pure black box.
Related trends appear in deep OD, where models such as explainable/robust autoencoders or prototype-based formulations seek to provide interpretable intermediate artifacts (e.g., representative patterns) alongside outlier decisions~\cite{angiulli2025explaining,robustexplainableae,binsPrototypes,prototypeAD,Zavrtanik_2021_ICCV}.
While attractive, such approaches are typically tied to a specific detector family and still follow the classical per-dataset training paradigm, which does not apply to tabular foundation models~\cite{XADSurvey1general,fomo}.

A second direction provides \emph{post-hoc} explanations for pre-trained detectors, including feature-importance approaches (e.g., Shapley-value based explanations for OD ensembles~\cite{deanshap}) and industrial root-cause style analyses built on feature relevance measures~\cite{FI4AD}.
Counterfactual explanations for outliers, sometimes framed as ``outlier repair'', aim to provide actionable guidance by proposing minimal changes that turn an outlier instance into a nominal one~\cite{counterfactual4AD}.
However, post-hoc explainers often incur substantial per-instance computation (sampling, optimization, repeated queries), and they typically explain the detector output rather than quantify whether the output itself is stable or reliable~\cite{XADSurvey2appl,XADSurvey1general}.

To the best of our knowledge, neither approach has been adapted for outlier foundation models. While post-hoc methods are technically applicable, we demonstrate a computationally highly attractive alternative. By training the foundation model to learn explanations directly, our approach provides accurate insights in negligible additional inference time, without compromising OD performance, enabling explanations that would otherwise be computationally infeasible to obtain.

\subsection{Uncertainty and Severity as Explainability Signals}

Uncertainty estimation can reveal epistemic instability, thereby identifying when predictions are unreliable.
In outlier detection (OD), uncertainty has therefore been explored as an auxiliary output complementing outlier scores, e.g., in uncertainty-aware time-series OD systems~\cite{uncertainAD2}.
For deep predictors, Monte Carlo dropout is a common operational proxy for epistemic uncertainty, but it increases test-time cost by requiring multiple stochastic forward passes~\cite{MCDropoutEpistemic}.
FoMo-X addresses this uncertainty-aware motivation while targeting single-pass deployment by distilling a dropout-based signal into a lightweight head evaluated once per query.

Severity signals, on the other hand, allow for prioritizing outliers by criticality rather than merely detecting their presence.
Recent work argues that standard outlier scores are often not severity-aligned and introduces a multilevel OD setting to create severity-aware outlier detectors~\cite{cao2024madbench}.

While uncertainty and severity are not feature-attribution explanations, they can provide \emph{diagnostic} artifacts for trust calibration and safe decision-making.
For instance, \cite{katzke2025skinsplain} recently proposed a framework to study how such cues can influence and calibrate user trust in a high-stakes domain.
Analogously, for OD, diagnostic signals can guide users on when to trust a flagged outlier versus when to escalate to deeper analysis or human review, which is particularly relevant for foundation models whose internal decision-making can otherwise be very difficult to inspect~\cite{XADSurvey1general,fomo}.

Conceptually, FoMo-X connects distilling expensive predictive procedures into cheaper predictors with probing learned representations using lightweight readouts.
Knowledge distillation compresses costly teachers into single models~\cite{hinton2015distillation}, and related dropout-distillation work approximates stochastic ensembling with reduced test-time cost~\cite{bulo2016dropoutdistill,gurau2018dropoutdistillation}.
Probing means training small heads on frozen representations to extract targeted attributes without modifying the backbone~\cite{alain2017probes}.
FoMo-X adopts these ideas in the OD foundation-model setting by training modular heads on frozen PFN embeddings with simulator-derived supervision, producing uncertainty- and severity-related diagnostics that would otherwise be either expensive or impossible to compute online while preserving the detector's original outputs~\cite{fomo}.

\section{Preliminaries}
\label{sec:background}

\subsection{Unsupervised Tabular Outlier Detection and PFNs}
\label{sec:problem-setting}

We consider tabular outlier detection (OD), where observations are feature vectors $x \in \mathbb{R}^d$ and the goal is to infer an outlier label $y \in \{0,1\}$, where $y=1$ denotes an outlier.
For the unsupervised setting, the established paradigm is to learn a dataset-specific detector by optimizing model parameters $\theta$ on a representative, predominantly normal, unlabeled training set. At inference, the trained model produces an outlier score $s_\theta(x)\in\mathbb{R}$  for a query $x$, and predicts $\hat y=\mathbb{I}[s_\theta(x)>\tau]$ for some threshold $\tau$.
However, a central practical bottleneck is that, without labels, selecting a suitable OD model family, hyperparameters, and a threshold is difficult and often dominates the end-to-end effort.

Prior-data fitted networks (PFNs) replace per-dataset training with a pretrained predictor that is capable of \emph{in-context learning (ICL)}. Given test query $x$ and dataset context $C$, a PFN can output a conditional outlier score $s_\theta(x;C)$
in a single forward pass, without gradient-based optimization of $\theta$ on the specific dataset.
Pretraining such a foundation model can be performed over a simulator-induced hypothesis space $\Phi$ as the data prior. Here, each hypothesis  $\phi\sim\Phi$  is specified by latent parameters and induces a joint distribution over dataset contexts and
labeled queries. A useful abstraction is that $\phi$ defines (i) an inlier-generating mechanism and (ii) an
outlier mechanism.
Training then minimizes the expected predictive loss under $\Phi$,
\begin{equation}
\label{eq:pfn-objective}
    \theta^\star \in \arg\min_\theta\;
    \mathbb{E}_{\phi\sim\Phi}\;
    \mathbb{E}_{(C,x,y)\sim p_\phi}\Big[\mathcal{L}\big(s_\theta(x;C),\, y\big)\Big].
\end{equation}

\subsection{FoMo-0D as a PFN for Zero-shot OD}
\label{sec:base-architecture}

FoMo-0D (Foundation Model for zero/0-shot OD)~\cite{fomo} is a transformer-based PFN specialized to the context-conditioned OD setting.
During inference, given an inlier-only training data context $C_{\mathrm{in}}$ and a test point $x\in C_{\mathrm{test}}$, it predicts an outlier score $s_\theta( x; C_{\mathrm{in}})$ in a single forward pass (zero-shot).
The pretraining is performed on a hypothesis space $\Phi$ based on Gaussian mixture models (GMMs), where a hypothesis $\phi \sim \Phi$ is specified by the dimensionality $d \in D$, number of components $m \in M$, centers $\{\mu_j\}^m_{j=1}\in [a,b]^d$ and covariances $\{\Sigma_j\}^m_{j=1} (diag(\Sigma_j) \in [a,b]^d)$ of a GMM, where the authors proposed $D=[5,100]$, $M=[1,5]$, $a=-5$ and $b=5$ as sensible default hyperparameters~\cite{fomo}.
Inliers are sampled from the GMM and retained if they lie within a $90th$ percentile region.
Outliers are synthesized from a variance-inflated variant that shares the same mixture centers but inflates the variances in a randomly drawn subset of dimensions, retaining only points outside the inlier percentile region.
Percentile membership is determined via Mahalanobis-distance thresholding using $\chi^2_d$ quantiles for each Gaussian mixture component.
This yields labeled synthetic datasets from which $C_{\mathrm{in}}$ (inlier-only) and $C_{\mathrm{test}}$ (mixed) are formed for labeled PFN training according to Equation~\ref{eq:pfn-objective}, with $\mathcal{L}$ being chosen as the binary cross-entropy over a 2-logit output (inlier vs outlier).
Internally, FoMo-0D encodes each sample as a token via a linear embedding layer, then applies a transformer encoder that enables sample-to-sample attention between context points and query-to-context attention for the test point, followed by an MLP readout head.

\section{FoMo-X}
\label{sec:fomox}

\begin{figure}[htbp]
    \centering
    \resizebox{\textwidth}{!}{
    \begin{tikzpicture}[
        node distance=0.5cm and 1cm,
        font=\sffamily\small,
        >={Latex[width=2mm,length=2mm]},
        block/.style={draw, fill=white, rectangle, minimum height=2.5em, minimum width=5em, rounded corners, drop shadow, align=center},
        input/.style={draw=none, fill=none, font=\bfseries, align=center},
        linear/.style={block, fill=blue!10, minimum width=5em},
        norm/.style={block, fill=green!10, minimum width=5em},
        transformer/.style={block, fill=orange!10, minimum height=8em, minimum width=8em, line width=1.2pt},
        gelu/.style={block, fill=yellow!10, minimum width=3.5em, minimum height=2em},
        arrow/.style={->, thick, draw=gray!80},
    ]

        \def\substr{0.6cm}

        \node[input] (input_tx) {Test Query\\(100 dim)};
        \node[input, below=1.8cm of input_tx] (input_x) {Inlier\\Context Data\\(100 dim)};

        \node[norm, right=0.8cm of input_tx] (norm_tx) {Normalize};
        \node[linear, right=0.8cm of norm_tx] (lin_tx) {Linear\\ \scriptsize 100 $\to$ 256};

        \node[norm] at (input_x -| norm_tx) (norm_x) {Normalize};
        \node[linear] at (input_x -| lin_tx) (lin_x) {Linear\\ \scriptsize 100 $\to$ 256};

        \node[transformer, right=1.2cm of lin_tx, yshift=-\substr] (te) {\textbf{Transformer}\\\textbf{Encoder}\\[0.5em]$\times$ 4 Layers};

        
        \node[linear, right=5cm of lin_tx] (h2_l1) {Linear\\ \scriptsize 256 $\to$ 512};
        \node[gelu, right=0.4cm of h2_l1] (h2_g) {GELU};
        \node[linear, right=0.4cm of h2_g] (h2_l2) {Linear\\ \scriptsize 512 $\to$ 2};
        \node[input, right=0.4cm of h2_l2] (out2) {Existing Outlier\\Detection Head};

        \node[linear, below=.7cm of h2_l1] (h1_l1) {Linear\\ \scriptsize 256 $\to$ 512};
        \node[gelu, right=0.4cm of h1_l1] (h1_g) {GELU};
        \node[linear, right=0.4cm of h1_g] (h1_l2) {Linear\\ \scriptsize 512 $\to$ 4};
        \node[input, right=0.4cm of h1_l2] (out1) {Severity Head};


        \node[linear, below=.7cm of h1_l1] (h3_l1) {Linear\\ \scriptsize 256 $\to$ 512};
        \node[gelu, right=0.4cm of h3_l1] (h3_g) {GELU};
        \node[linear, right=0.4cm of h3_g] (h3_l2) {Linear\\ \scriptsize 512 $\to$ 1};
        \node[input, right=0.4cm of h3_l2] (out3) {Uncertainty Head};

        
        \draw[arrow] (input_tx) -- (norm_tx);
        \draw[arrow] (norm_tx) -- (lin_tx);
        \draw[arrow] (lin_tx) -- ([yshift=\substr]te.west); 


        \draw[arrow] (input_x) -- (norm_x);
        \draw[arrow] (norm_x) -- (lin_x);
        \draw[arrow] (lin_x.east) -| ([xshift=-0.5cm]te.south);

        \coordinate (split) at ($([yshift=\substr]te.east)!0.4!(h2_l1.west)$);
        \draw[thick, draw=gray!80] ([yshift=\substr]te.east) -- (split);
        \draw[arrow] (split) |- (h1_l1.west);
        \draw[arrow] (split) -- (h2_l1.west);
        \draw[arrow] (split) |- (h3_l1.west);

        \draw[arrow] (h1_l1) -- (h1_g); \draw[arrow] (h1_g) -- (h1_l2); \draw[arrow] (h1_l2) -- (out1);
        \draw[arrow] (h2_l1) -- (h2_g); \draw[arrow] (h2_g) -- (h2_l2); \draw[arrow] (h2_l2) -- (out2);
        \draw[arrow] (h3_l1) -- (h3_g); \draw[arrow] (h3_g) -- (h3_l2); \draw[arrow] (h3_l2) -- (out3);

    \end{tikzpicture}
    }
    \caption{Architecture of FoMo-X.}
    \label{fig:architecture}
\end{figure}
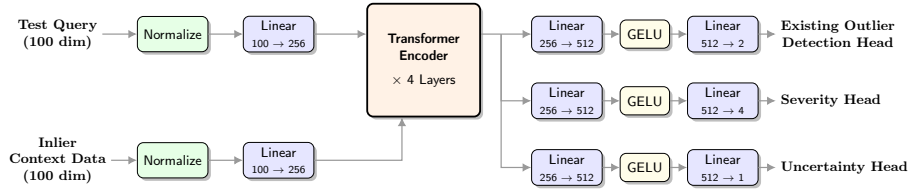

We now introduce \textbf{FoMo-X} (\underline{F}ear \underline{O}f \underline{M}issing \underline{O}utlier e\underline{X}planations), a modular framework that augments a pretrained outlier PFN ---in our case FoMo-0D~\cite{fomo}--- with additional explainability signals. FoMo-X is designed as an add-on that attaches lightweight auxiliary heads to the embedding computed for each query point, yielding low-overhead, per-sample diagnostic information without changing the underlying outlier-detection predictions.

Namely, while FoMo-0D outputs only an outlier prediction per query sample, it internally computes a rich, context-dependent representation that encodes how the query relates to the context-defined normality.
In turn, we can decompose this output into an encoder $E_{\theta}$ and an outlier detection head:
\begin{equation}
\label{eq:fomo-decomp}
    z \;=\; E_{\theta}(x;C) \in \mathbb{R}^{d_z},
    \qquad
    s_{\theta}( x;C) \;=\; h^{\mathrm{OD}}_{\theta_0}(z) \in [0,1],
\end{equation}
where $z$ denotes the context-conditioned query embedding and $h^{\mathrm{OD}}_{\theta_0}$ maps to the outlier prediction. We refer to $z$ as the \emph{FoMo embedding} of $x$ given context $C$.
FoMo-X introduces a set of $K$ auxiliary heads
\begin{equation}
    h^{(k)}_{\theta_k}:\mathbb{R}^{d_z} \rightarrow \mathcal{E}_k,\qquad k\in\{1,\dots,K\},
\end{equation}
which produce additional diagnostic explainability signals $\hat e_k =  h^{(k)}_{\theta_k}\!\left(z\right).$

A key design requirement is that FoMo-X must not alter the outlier detector’s behavior. We therefore freeze the pretrained parameters of both $E_{\theta}$ and $h^{\mathrm{OD}}_{\theta_0}$ and train only the auxiliary parameters $\{\theta_k\}_{k=1}^K$. Since each auxiliary head reads out $z$ but does not feed information back into the backbone, the outlier prediction remains identical.
While any lightweight readout operating on $z$ is valid, in our instantiation, we use small MLP heads (Figure~\ref{fig:architecture}) that mirror the original FoMo-0D classifier head design (linear projection, nonlinearity, output layer with a hidden dimension of $512$), so that auxiliary outputs incur negligible overhead relative to the transformer encoder.

\subsection{Training Auxiliary Heads}
\label{sec:train-heads}

\begin{figure}[t]
    \centering
    \includegraphics[width=1.0\linewidth]{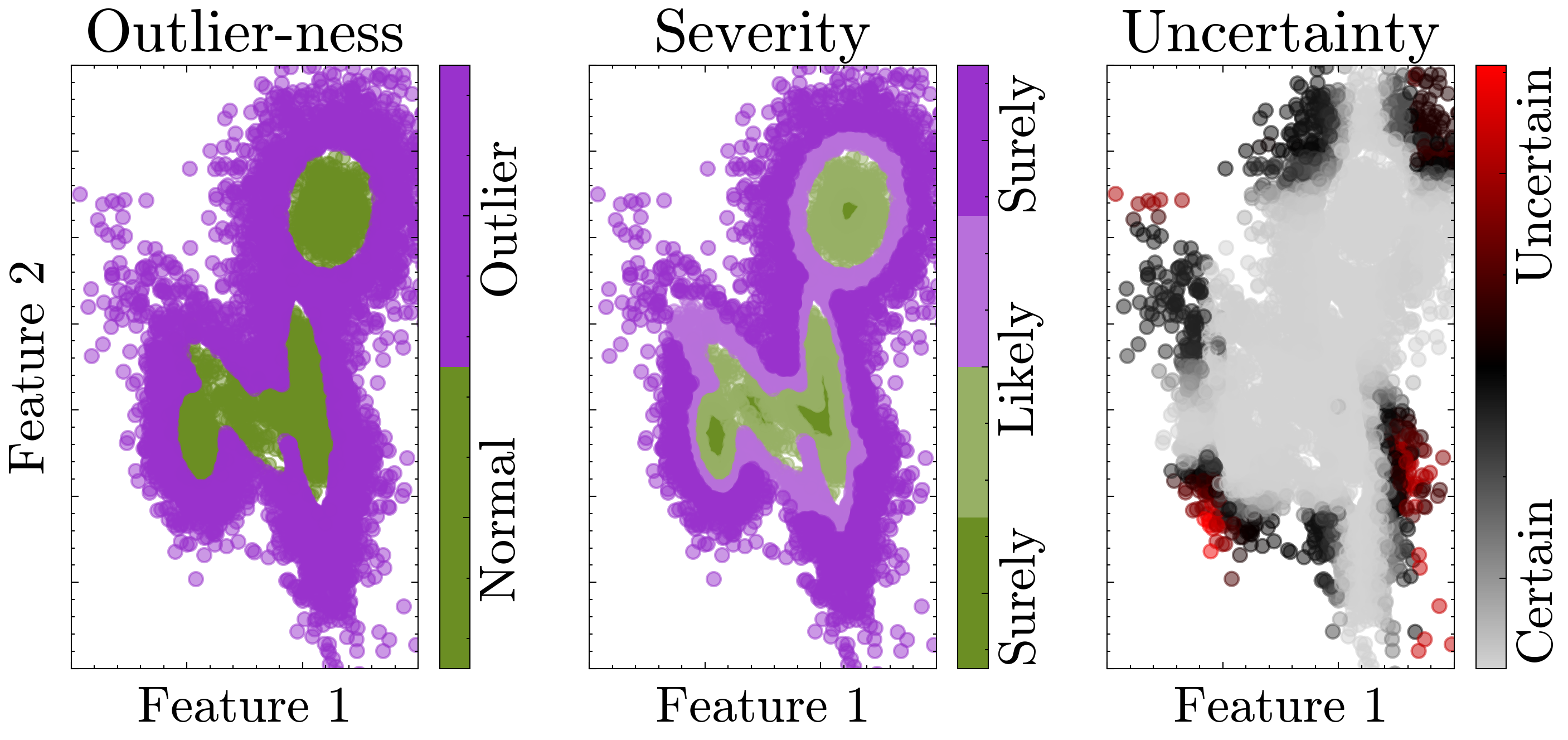}
    \caption{Example of a \textbf{synthetic context dataset used to train FoMo-X}. The left plot indicates whether a sample is an outlier and corresponds to the supervision used for the original FoMo-0D head, whereas the remaining plots show the ground-truth targets used to train the auxiliary heads. The middle plot shows outlier severity, and the right plot shows epistemic uncertainty under the FoMo-0D model.}
    \label{fig:example}
\end{figure}

FoMo-X heads are trained offline using automatically generated supervision from synthetic tasks drawn from a distribution induced by a simulator over OD tasks. Concretely, let $\Phi$ denote the simulator-induced distribution used to generate head-training contexts and queries. For each head $k$, we define a task- and sample-dependent target generation function
\begin{equation}
    g_k:\; (\phi,C,x,y)\mapsto e_k \in \mathcal{E}_k,
\end{equation}
which produces the desired diagnostic label $e_k$ for a query $x$ given context $C$ with the outlier ground truth $y$ induced under hypothesis $\phi$. Target generation can depend on quantities that are available in simulation (e.g., likelihood under the generating distribution) or on an offline probe that is too expensive for deployment (e.g., Monte Carlo dropout). In all cases, target computation is performed on simulated data during head training only.

Each auxiliary head is trained on frozen embeddings by minimizing an expected loss under the task prior:
\begin{equation}
    \theta_k^\star \in
    \arg\min_{\theta_k}\;
    \mathbb{E}_{\phi\sim\Phi}\;
    \mathbb{E}_{(C,x,y)\sim p_\phi}\Big[
        \mathcal{L}_k\!\big(
            h^{(k)}_{\theta_k}(E_{\theta}(x;C)),\;
            g_k(\phi,C,x,y)
        \big)
    \Big],
\end{equation}
where $\mathcal{L}_k$ is a head-specific loss.
Because the backbone is fixed, heads can be trained independently and added iteratively. This supports a workflow in which a deployment may start with the base detector to later be augmented with additional explainability signals, without retraining FoMo-0D or previously trained heads.
In the remainder of this section, we instantiate FoMo-X with two heads: a \emph{severity} head providing a four-level significance tiering of the outlier decision and an \emph{uncertainty} head distilling dropout-based epistemic instability, see Figure~\ref{fig:architecture}. Before formalizing them, Figure~\ref{fig:example} provides an example for ground truth as specified by their respective target generation functions.

 \subsection{Instantiation I: Severity Head}
\label{sec:severity-head}

FoMo-0D is trained as a binary classifier that separates normal and outlier samples given context. However, under the simulator that generates training tasks, outlier labels arise from a continuous notion of normality: samples can be \emph{more} or \emph{less} likely under the generating distribution. FoMo-X leverages this structure to define an interpretable severity tiering: points that are clearly either normal or abnormal should be distinguished from those closer to the decision boundary between the two.

Consider a simulated hypothesis $\phi\sim\Phi$ with an inlier GMM specified by mixture weights
$\{\pi_j\}_{j=1}^{m}$, component means $\{\mu_j\}_{j=1}^{m}$ and covariances
$\{\Sigma_j\}_{j=1}^{m}$. In our simulator, each hypothesis additionally specifies a (latent) subset of ``active'' features $S_\phi\subseteq\{1,\dots,d\}$ s.t. all density-based quantities below are computed on the projected vector $x_{S_\phi}$.
Namely, as a continuous notion of normality, we use the maximum component log-density
\begin{equation}
    r_\phi(x)
    \;:=\;
    \max_{j\in[m]}
    \log \mathcal{N}\!\big(x_{S_\phi}\,\mid\,\mu_{j,S_\phi},\,\Sigma_{j,S_\phi}\big),
\end{equation}
where larger values indicate that $x$ lies in a higher-density region of the inlier mixture.
Let $y\in\{0,1\}$ denote the simulator-induced inlier/outlier label.

To obtain a four-class severity target, we split each class into ``sure'' and ``likely'' outlier and normal regions using
within-class medians of $r_\phi(x)$:
\begin{equation}
    m_0 := \mathrm{median}\{r_\phi(x): y=0\},\qquad
    m_1 := \mathrm{median}\{r_\phi(x): y=1\}.
\end{equation}
We then define the severity tier $e_{\mathrm{sev}}\in\mathcal{E}_{\mathrm{sev}}$ with
$\mathcal{E}_{\mathrm{sev}}=\{\mathrm{SN},\mathrm{LN},\mathrm{LO},\mathrm{SO}\}$ via
\begin{equation}
    g_{\mathrm{sev}}(\phi,C,x,y)=
    \begin{cases}
        \mathrm{SN} & \text{if } y=0 \text{ and } r_\phi(x)> m_0,\\
        \mathrm{LN} & \text{if } y=0 \text{ and } r_\phi(x)\le m_0,\\
        \mathrm{LO} & \text{if } y=1 \text{ and } r_\phi(x)> m_1,\\
        \mathrm{SO} & \text{if } y=1 \text{ and } r_\phi(x)\le m_1.\\
    \end{cases}
\end{equation}

The severity head maps the FoMo-0D embedding $z$ to logits
$h^{\mathrm{sev}}_{\theta_{\mathrm{sev}}}(z)\in\mathbb{R}^{4}$ and is trained via cross-entropy against the one-hot encoded target $g_{\mathrm{sev}}(\phi,C,x,y)$:
\begin{equation}
    \mathcal{L}_{\mathrm{sev}}
    \;=\;
    -\, g_{\mathrm{sev}}(\phi,C,x,y)^{\top}\!
    \log \mathrm{softmax}\!\left(h^{\mathrm{sev}}_{\theta_{\mathrm{sev}}}(z)\right).
\end{equation}

On real-world datasets, the ground-truth likelihood under the generating simulator is not available. We therefore interpret predicted severity as an intrinsic reliability tier: ``sure'' predictions are expected to be more stable and less error-prone than ``likely'' ones, conditional on the model’s own internal representation.

\subsection{Instantiation II: Uncertainty Head}
\label{sec:uncertainty-head}

For transformer-based predictors, a commonly used operational proxy for epistemic uncertainty is Monte Carlo dropout: repeating stochastic forward passes with dropout enabled and measuring predictive variability. While informative, this procedure increases inference cost substantially if executed online. We therefore distill a dropout-based uncertainty estimate into a single auxiliary head evaluated in one forward pass.

Let $p_t  = s_{\theta_0}(x;C,\xi)\in[0,1]$ denote the outlier probability of the frozen detector under a sampled dropout mask $\xi$ (a stochastic forward pass with dropout enabled). For an integer $M\ge2$, we draw i.i.d. masks $\xi_1,\dots,\xi_M$ and compute
\begin{equation}
    u(x;C)
    \;:=\;
    \mathrm{Std}\{p_t\}_{t=1}^{M}
    \;=\;
    \sqrt{\frac{1}{M-1}\sum_{t=1}^{M}(p_t-\bar p)^2},
    \qquad
    \bar p \;=\; \frac{1}{M}\sum_{t=1}^{M} p_t.
\end{equation}
To combat heteroscedasticity and to increase numerical stability, we use the log-uncertainty as the regression target:
\begin{equation}
    g_{\mathrm{unc}}(\phi,C,x)\;:=\;\log\big(u(x;C)+\varepsilon\big),
\end{equation}
with a small $\varepsilon > 0$ and $e_{\mathrm{unc}}\in\mathcal{E}_{\mathrm{unc}}=\mathbb{R}$. Note that $g_{\mathrm{unc}}$ does not depend explicitly on $\phi$ beyond providing $(C,x)$.
The uncertainty head predicts a scalar and is trained via a regression loss, namely, mean absolute error:
\begin{equation}
    \mathcal{L}_{\mathrm{unc}}
    \;=\;
    \Big| h^{\mathrm{unc}}_{\theta_{\mathrm{unc}}}(z) - g_{\mathrm{unc}}(\phi,C,x)\Big|.
\end{equation}

At deployment, the uncertainty head provides a one-pass approximation to the dropout-based variability proxy, enabling uncertainty-aware prediction stability assessment and diagnostics without repeated stochastic forward passes.

\section{Experiments}

We empirically assess FoMo-X along three questions: (i) fidelity under the simulator prior: can auxiliary heads recover their simulator-defined targets when trained on frozen FoMo-0D embeddings? (ii) utility and transfer: do the learned diagnostics provide meaningful signals on real-world datasets, despite being trained purely on synthetic tasks? (iii) efficiency: what is the additional computational overhead of producing these diagnostics at inference time?

\subsection{Experimental Setup}

We build on the pretrained FoMo-0D outlier detection foundation model and freeze all backbone and original OD-head parameters throughout.
All auxiliary heads follow the architecture in Figure~\ref{fig:architecture} (a lightweight MLP readout on the frozen query embedding), ensuring that the original outlier scores remain unchanged for any input.

To train the severity and uncertainty heads, we generate synthetic OD tasks from a GMM-based simulator distribution that closely follows the FoMo-0D training prior, with minor adaptations.\footnote{
We sample the dimensionality uniformly, $d \sim \mathrm{Unif}\{5,\dots,100\}$, and the number of mixture components uniformly up to five, $m \sim \mathrm{Unif}\{1,\dots,5\}$. Mixture means are drawn i.i.d.\ from $\mathrm{Unif}([-5,5]^d)$. In contrast to FoMo-0D's diagonal-only covariance restriction, we use \emph{full} (non-diagonal) positive-definite covariances by sampling random eigenbases and \emph{exponentially distributed} eigenvalue scales (to cover multiple feature scales). Each synthetic dataset contains 5000 inlier and 5000 outlier samples; a random subset is used as inlier-only context and as test queries. Outliers correspond to low-density samples under the inlier-generating distribution (i.e., sampled from the complement of a high-density region), enabling supervision for the head targets.
}
For each synthetic task, we compute simulator-derived targets as described in Sections~\ref{sec:severity-head}--\ref{sec:uncertainty-head}: a four-level severity label and a dropout-based uncertainty target.
Each head is trained independently on frozen embeddings for 200 epochs. Each epoch consists of 250 batches with 8 synthetic datasets per batch, i.e., $250 \cdot 8 = 2000$ synthetic datasets per epoch and 400k datasets per head in total.
This amounts to 25\% of the number of datasets used during the original FoMo-0D pretraining~\cite{fomo}.
We optimize using Adam with an initial learning rate of 0.001, decayed by 20\% every 10 epochs down to a final learning rate of $1.15 \cdot 10^{-5}$.
For the uncertainty head, we compute the Monte Carlo dropout uncertainty target using $M=10$ stochastic forward passes with dropout enabled at probability $p=0.1$, and we set $\varepsilon=10^{-6}$. Both heads use the same MLP architecture as depicted in Figure~\ref{fig:architecture}.

To evaluate transfer beyond the simulator prior, we use the native tabular subset of ADBench~\cite{surveyzhao}, consisting of 47 datasets from diverse domains.
We restrict to this subset as it mostly matches FoMo-0D's tabular input regime (up to 100 features).
For each dataset, FoMo-0D receives an inlier-only context $C_{\mathrm{in}}$ and predicts outlier scores for the test points. Labels are used \emph{only} for evaluation (e.g., AUROC reporting and outlier-fraction analyses).

\begin{figure}[t]
    \centering
    \includegraphics[width=1.0\linewidth]{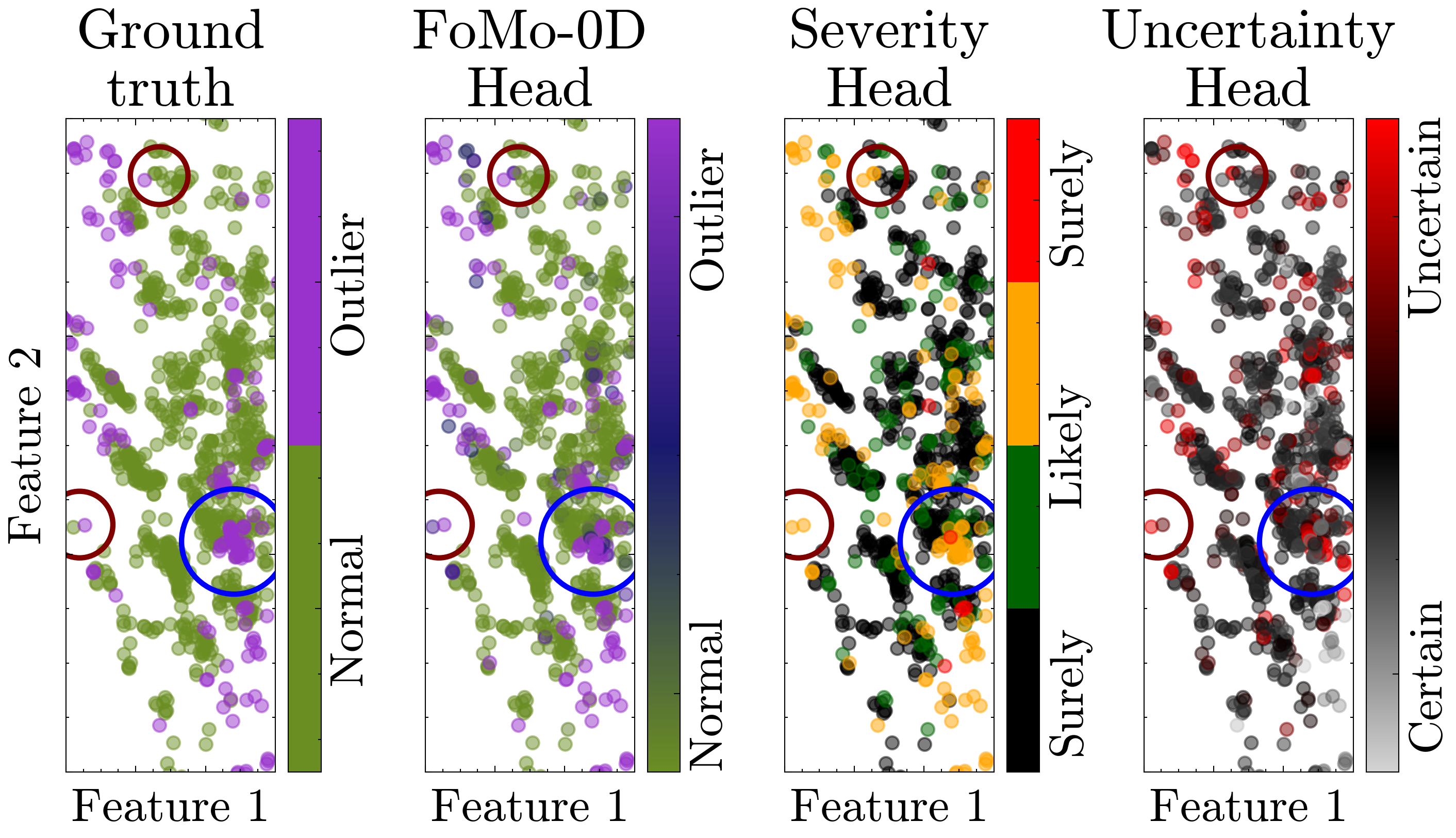}
    \caption{FoMo-X explanations for the cardio dataset, visualized using PCA~\cite{pcaAlgo}. The red circles highlight prediction errors, corresponding to differences between the ground truth (center left) and the outlier detection head (left). These errors generally occur in regions of low severity (center right) and high uncertainty (right). The blue circle illustrates how severity can help interpret a complex decision boundary.}
    \label{fig:one}
\end{figure}

Figure~\ref{fig:one} provides a qualitative illustration on the ADBench \texttt{cardio} dataset (ID \texttt{6\_cardio}), chosen because it is low-dimensional enough for PCA visualization and yields a non-trivial detection quality (FoMo-0D AUROC of 0.88).
The plot highlights two desirable behaviors: (i) prediction errors (red circles) tend to occur near ambiguous regions where the severity head outputs ``likely'' tiers and the uncertainty head assigns high uncertainty; and (ii) severity tiers provide a coarse view of the decision boundary structure (blue circle), complementing the scalar, almost binary, outlier prediction.

For the severity head, we report balanced accuracy~\cite{balancedAccuracy} on synthetic validation data (four-class classification)\footnote{
We use balanced accuracy because class frequencies across severity tiers are not guaranteed to be uniform; standard accuracy yields similar trends in our setting.
}. On ADBench, no ground-truth severity exists; we therefore evaluate the \emph{utility} of severity tiers as an intrinsic reliability signal through error rates conditioned on predicted tiers (Table~\ref{tab:confusion_matrix}) and outlier fractions per predicted tier across datasets (Figure~\ref{fig:severity}).
We evaluate the uncertainty head using Spearman rank correlation $\rho_s$ between predicted uncertainty and the MC-dropout teacher uncertainty on the same dataset~\cite{SpearmanCorrCompare}. We report correlations per dataset and summarize their distribution across datasets (Figure~\ref{fig:corr_variance}).

\subsection{Training Dynamics}
\label{sec:training_dynamics}

Both heads were trained on a single Nvidia RTX Pro Blackwell GPU. Training required 12 hours for the severity head and 18 hours for the uncertainty head. Synthetic data generation was performed on an Intel Xeon w9-3495X CPU (56 physical cores). These costs are offline and incurred only once per head.

\begin{figure}[t]
    \centering
    \includegraphics[width=0.7\linewidth]{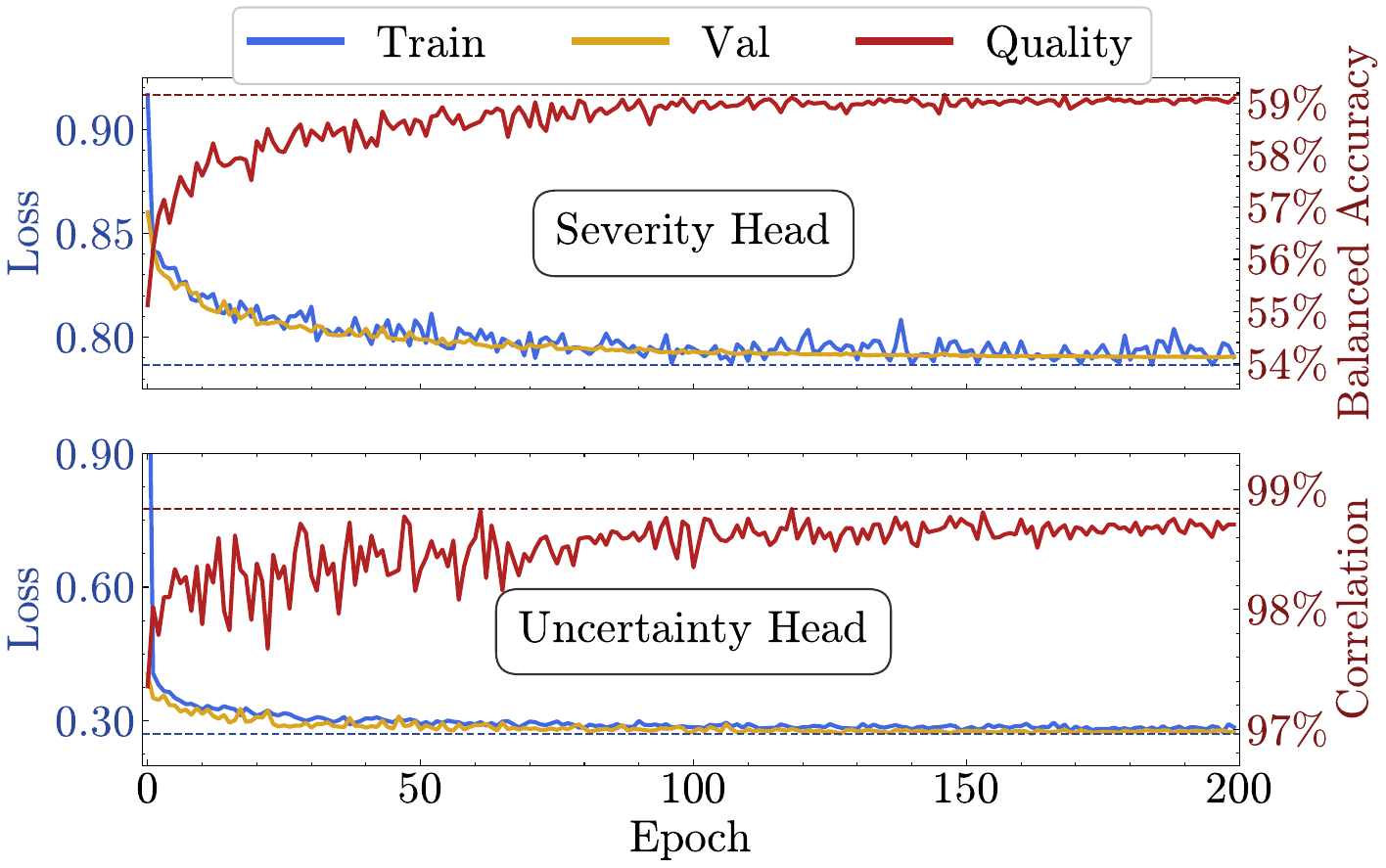}
    \caption{Training progress of both proposed heads. While performance continues to improve throughout training, \textbf{both heads already achieve high accuracy after only a few epochs}.}
    \label{fig:progress}
\end{figure}

Figure~\ref{fig:progress} summarizes training loss and validation quality over epochs.
Because the FoMo-0D backbone remains frozen, optimization is stable and converges rapidly, consistent with the interpretation that both diagnostic targets are already strongly encoded in the pretrained query embedding.

After a single epoch, the uncertainty head already achieves a Spearman correlation of 97\% on validation tasks, and improves to nearly 99\% after full training. The severity head reaches 55\% balanced accuracy after one epoch and 59\% after 200 epochs.
While this implies that we can train additional heads in significantly fewer epochs, we train for the full training timeline and report the results to emphasize the full potential of our approach.

\subsection{Results: Severity Head}
\label{sec:severity_results}

\begin{table}[b!]
\caption{Confusion matrix of our severity predictions for the cardio dataset. \textbf{The model makes fewer mistakes when it is sure of its prediction.}}
\vspace{0.2em}
\centering
\begin{tabular}{lccc}
\hline
\textbf{Prediction} & \textbf{True Normal} & \textbf{\hspace{0.6em}True Outlier\hspace{0.6em}}  & \textbf{Fraction of Mistakes} \\
\hline
Surely Normal      & $684$ & $4$  &  $0.58\%$  \\
Likely Normal      & $112$ & $20$ &  $15.15\%$  \\\hline
Likely Outlier   & $31$  & $146$ & $17.51\%$  \\
Surely Outlier   & $1$   & $6$   & $14.29\%$  \\
\hline
\end{tabular}
\label{tab:confusion_matrix}
\end{table}

On synthetic validation tasks, the severity head achieves 59\% balanced accuracy (four-way classification), substantially above the 25\% expected from a trivial classifier.
We do not expect perfect accuracy here: the simulator defines ``sure'' vs.\ ``likely'' boundaries via within-class medians of a latent density score, which induces inherent ambiguity and cannot always be recovered from finite context alone.
Nevertheless, the head recovers a meaningful tiering signal that can be used as an intrinsic reliability indicator.

Since ADBench does not provide ground-truth severity labels, we evaluate whether the predicted severity tiers correlate with \emph{decision reliability}.
Table~\ref{tab:confusion_matrix} reports the confusion matrix for \texttt{cardio} when collapsing severity tiers into normal vs.\ outlier decisions (SN/LN vs.\ LO/SO). The mistake rates are sharply tier-dependent: ``Surely Normal'' has only 0.58\% mistakes, whereas ``Likely Normal'' has 15.15\% mistakes, indicating that the model's internal ``sure''/``likely'' split provides actionable reliability information. 

\begin{figure}[t!]
    \centering
    \begin{subfigure}[b]{0.45\linewidth}
        \centering
        \includegraphics[width=\linewidth]{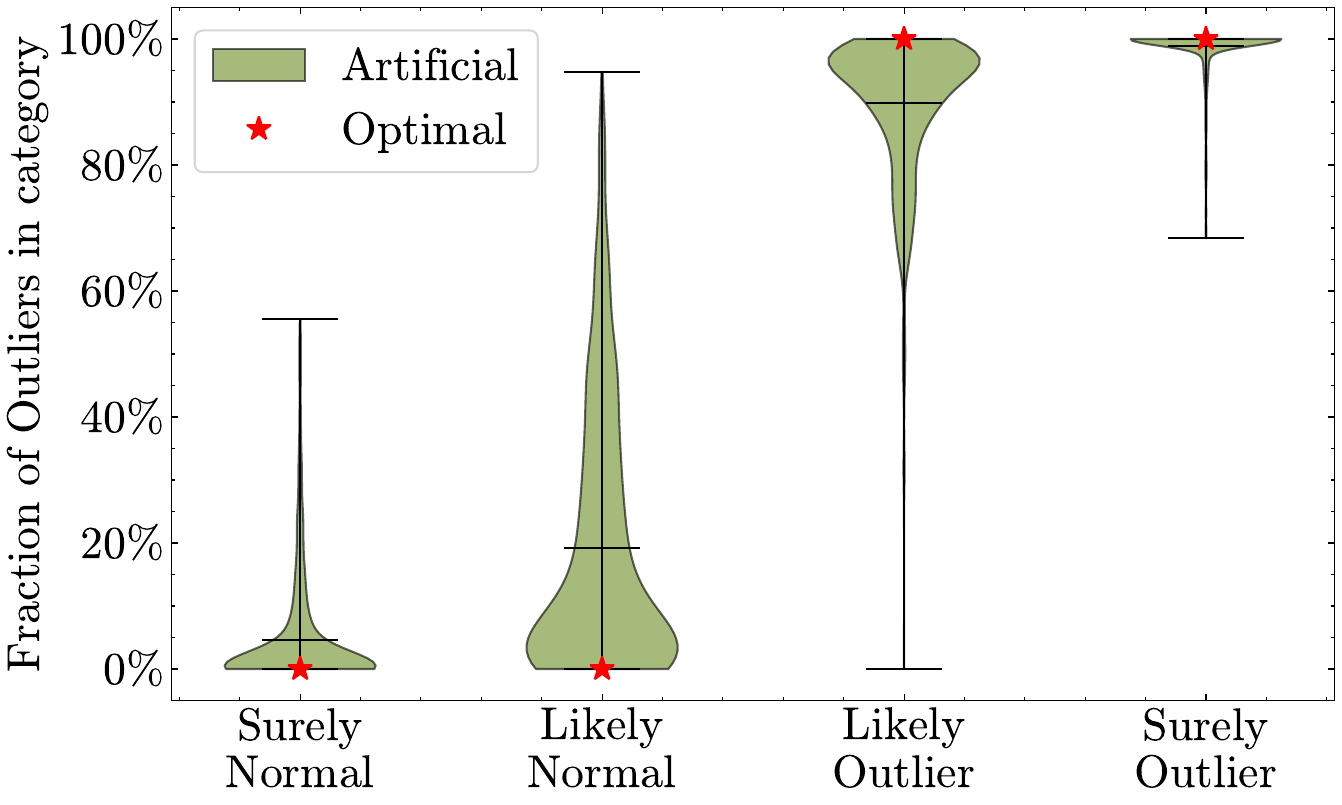}
        \caption{Fraction of outliers by severity head prediction. We generate $1000$ artificial test datasets, and show the distribution of outlier fractions for each of these in a violin plot.}
       
        \label{fig:asev}
    \end{subfigure}
    \hfill
    \begin{subfigure}[b]{0.45\linewidth}
        \centering
        \includegraphics[width=\linewidth]{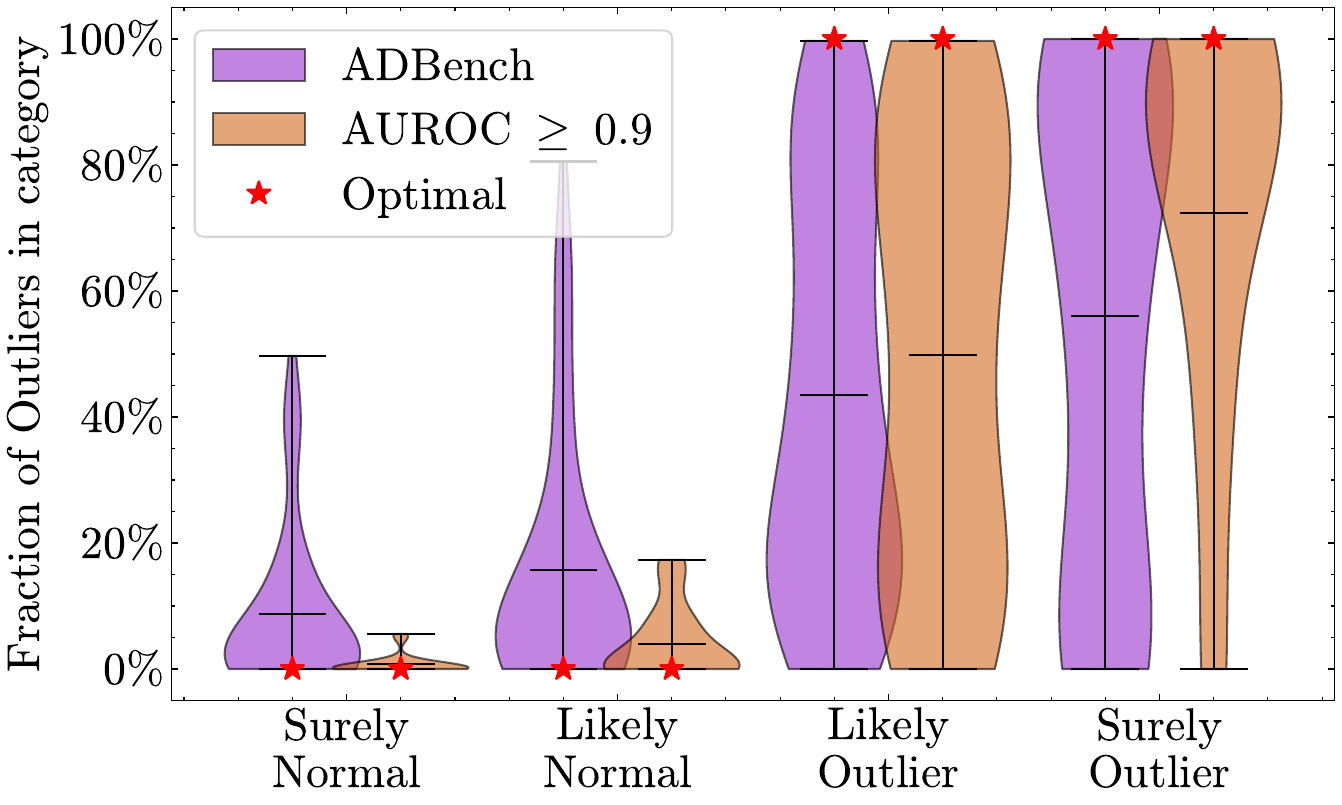}
        \caption{Fraction of outliers for real-world datasets. We give the distribution for all datasets in purple, and only those datasets that are well separable by FoMo-0D in orange.}
        \label{fig:sev}
    \end{subfigure}
    \caption{\textbf{Predictions of ``surely'' by the severity head are more reliable than predictions of ``likely''.} This pattern is observed for artificial test data (left) and also generalizes to real-world datasets (right).}
    \label{fig:severity}
\end{figure}

This behavior generalizes beyond a single dataset. Figure~\ref{fig:asev} shows, on 1000 synthetic test datasets, that predicted ``surely'' categories are substantially cleaner: ``Surely Normal'' contains near-zero outliers, and ``Surely Outlier'' contains near-all outliers, whereas the ``likely'' tiers concentrate the ambiguous region.
Figure~\ref{fig:sev} shows the same outlier-fraction stratification on ADBench. The separation is clearest on datasets where FoMo-0D is already accurate (AUROC $\ge 0.9$), which is expected: when the base detector is weak or labels are noisy, any internal tiering signal becomes less distinct.
Finally, Figure~\ref{fig:one} qualitatively illustrates the same phenomenon: FoMo-0D's misclassifications concentrate in regions where the severity head assigns ``likely'' tiers (decision-boundary proximity), supporting the interpretation of severity as an intrinsic reliability tier rather than as an externally validated ground-truth notion of harm.

\subsection{Results: Uncertainty Head}
\label{sec:uncertainty_results}

\begin{figure}[t]
    \centering
    \begin{subfigure}[b]{0.45\linewidth}
        \centering
        \includegraphics[width=\linewidth]{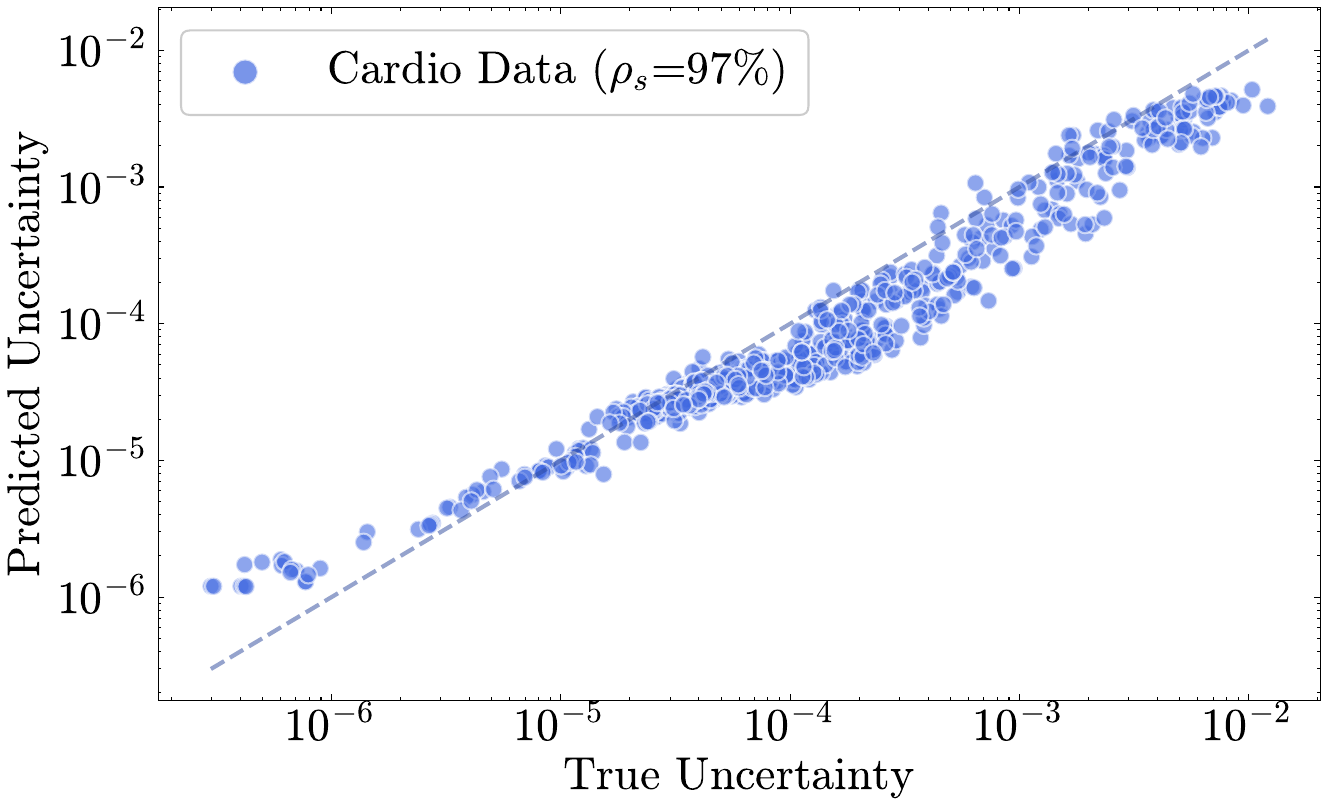}
        \caption{Example for cardio dataset.}
        \label{fig:aunc}
    \end{subfigure}
    \hfill
    \begin{subfigure}[b]{0.45\linewidth}
        \centering
        \includegraphics[width=\linewidth]{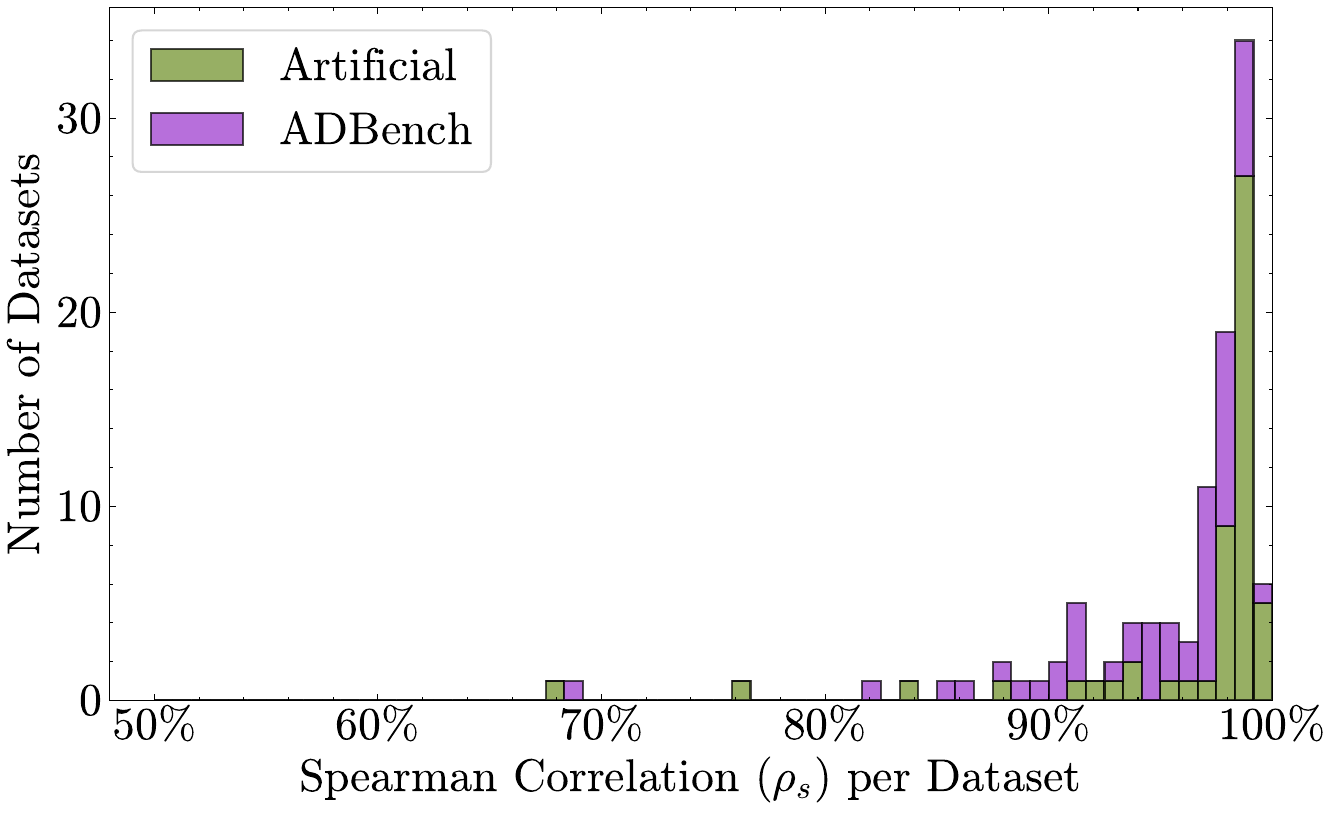}
        \caption{Overview across datasets.}
        \label{fig:unc}
    \end{subfigure}
    \caption{The uncertainty head generalizes to real-world data: We show an illustrative comparison for the cardio dataset in blue, along with a histogram of Spearman correlations across artificial test data (green) and ADBench datasets (purple). \textbf{For nearly all datasets, the correlation between ground-truth and predicted uncertainty exceeds 90\%.}}
    \label{fig:corr_variance}
\end{figure}

On synthetic validation tasks, the uncertainty head achieves near-perfect rank agreement with the MC-dropout teacher (up to $\approx$99\% Spearman correlation; Figure~\ref{fig:progress}).
To evaluate transfer, we compute the same MC-dropout uncertainty on ADBench and compare it to the head prediction.
Figure~\ref{fig:aunc} shows an example scatter plot on \texttt{cardio} with $\rho_s=97\%$, indicating that the head preserves uncertainty rankings closely over several orders of magnitude and validating the predicted uncertainties in Figure~\ref{fig:one}.
Across all ADBench datasets, Figure~\ref{fig:unc} shows that most datasets achieve $\rho_s \ge 90\%$, demonstrating that the distilled single-pass head generalizes well despite being trained purely on synthetic tasks.
Operationally, this provides a practical approximation to dropout-based epistemic instability without repeated stochastic forward passes.

\subsection{Runtime}
\label{sec:runtime}

\begin{figure}[t]
    \centering
    \includegraphics[width=0.7\linewidth]{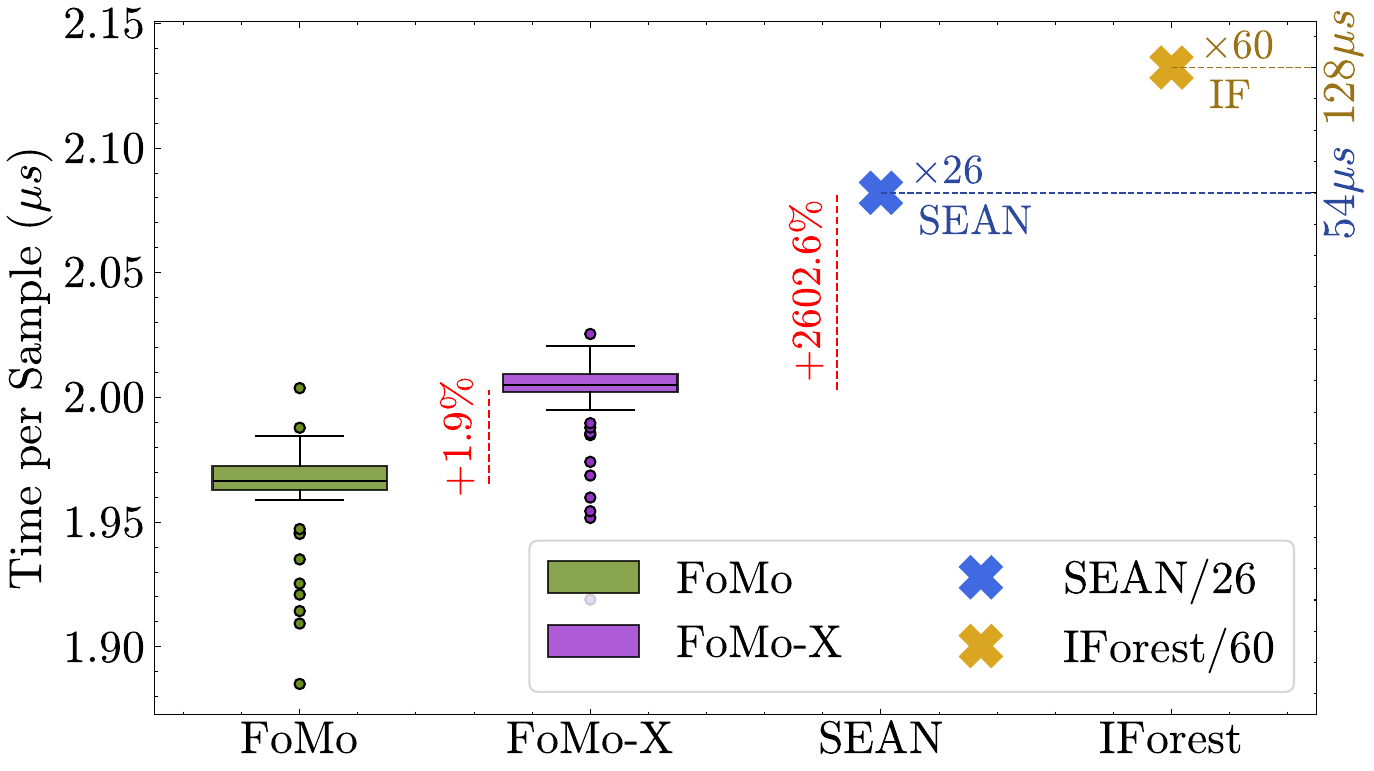}
    \caption{\textbf{FoMo-X explanations incur negligible computational overhead}: Combined training and inference time of FoMo and FoMo-X (zero-shot) compared with that of the fastest algorithms from Reference~\cite{sean} on ADBench. The additional cost of explanation generation is below $1\%$ per head and remains negligible relative to even the fastest competitors.}
    \label{fig:time}
\end{figure}

Foundation models are often limited by the computational resources required to run them. Thus, we also evaluate the computational overhead incurred by FoMo-X at inference. 

Following the FoMo-0D evaluation protocol~\cite{fomo}, we report average prediction time per test sample on ADBench by predicting each dataset 100 times.
Figure~\ref{fig:time} shows that our implementation requires less than 2 $\mu$s per sample for FoMo-0D inference, and adding \emph{both} auxiliary heads increases this time by only 36 ns ($<2\%$ overhead; $<1\%$ per head).
To contextualize these costs, we also compare to reported runtimes of highly optimized fast OD baselines from~\cite{sean}, which still remain more than $27\times$ slower per sample in this setting.

Absolute runtimes depend on implementation details (e.g., batching and hardware).\footnote{
The original FoMo-0D paper reports 7.7 ms per sample; our substantially lower absolute time is largely due to batch-wise GPU inference in our implementation. We therefore emphasize relative overheads (FoMo vs.\ FoMo-X) as the main takeaway for deployment.
}
However, importantly, the marginal cost of computing the auxiliary diagnostics is negligible relative to the base model and especially to competitors, leaving room for additional heads in future work.

\section{Discussion and Limitations}\label{sec:discuss}

FoMo-X demonstrates that a frozen OD PFN can be augmented with
lightweight diagnostic heads that are evaluated in a single forward pass, adding
operationally useful signals (severity tiers and uncertainty) at negligible runtime
overhead. At the same time, our ablations indicate that the feasibility and
transferability of additional heads depends strongly on the type of target being
predicted and on the degree of mismatch between the simulator-induced training
prior and real-world datasets. In the following, we (i) discuss observed failure
modes when training further heads beyond severity and uncertainty, and (ii)
summarize key limitations of current OD foundation model priors and
architectures that constrain generalization and the space of explainability
signals that can be exposed.

\subsection{Training Further Heads}\label{sec:limit}

\begin{figure}[t]
    \centering
    \begin{subfigure}[b]{0.45\linewidth}
        \centering
        \includegraphics[width=\linewidth]{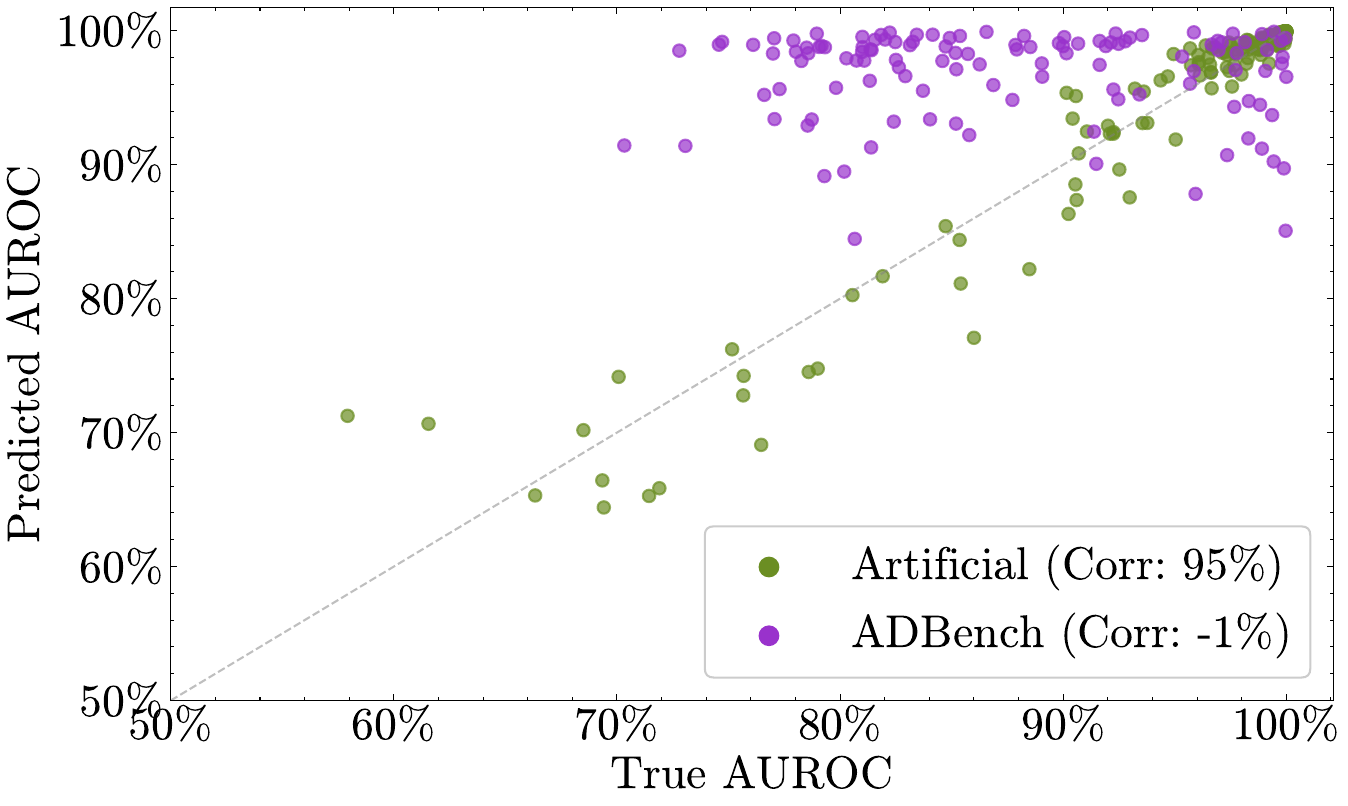}
        \caption{Predicting the outlier detection performance. On artificial data, this head is very effective, whereas for real world data, the head only predicts noise.}
        \label{fig:rocauc}
    \end{subfigure}
    \hfill
    \begin{subfigure}[b]{0.45\linewidth}
        \centering
        \includegraphics[width=\linewidth]{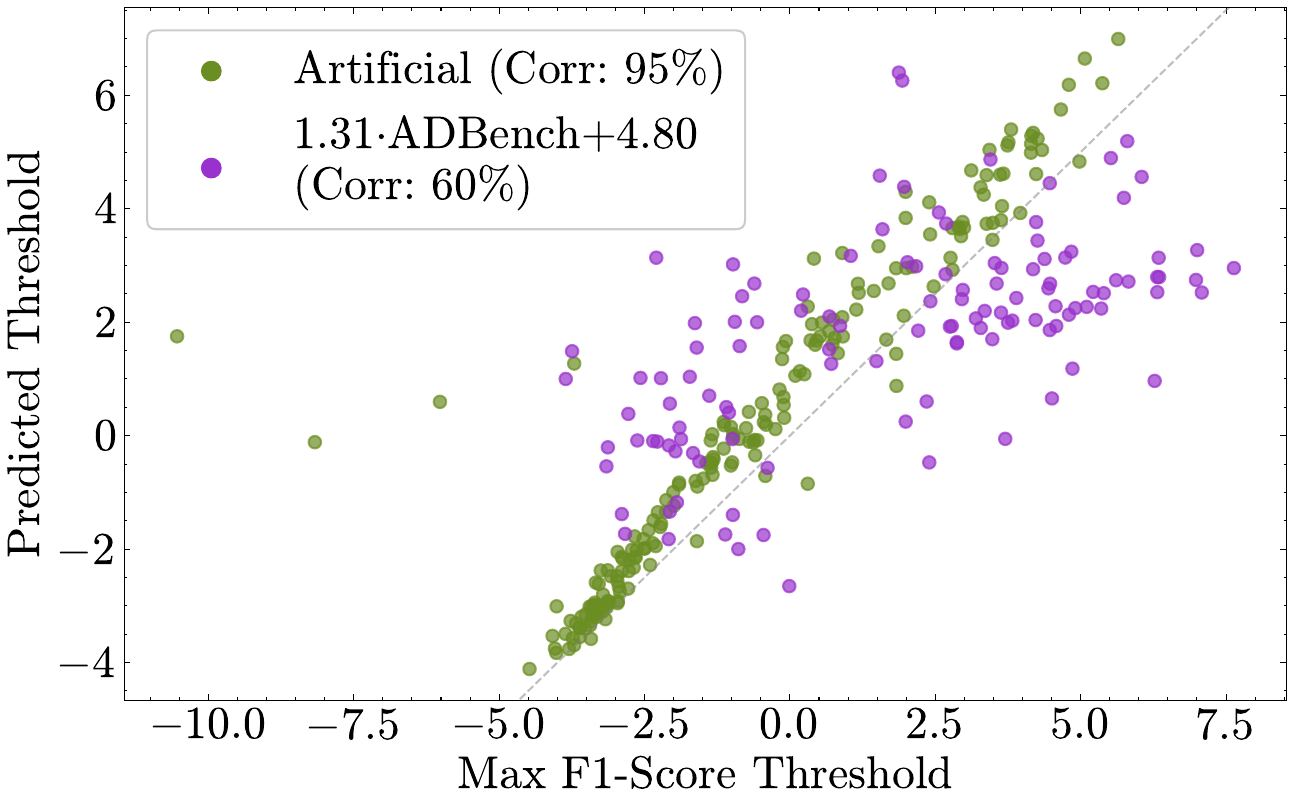}
        \caption{Maximum F1-score thresholds predicted compared to ground truth. Here shown with a linear correction ($y\rightarrow1.31\cdot y+4.8$), not affecting correlation.}
        \label{fig:thresh}
    \end{subfigure}
    \caption{\textbf{Some heads show weaker generalization from artificial to real-world data:} One head is trained to predict the outlier detection performance of FoMo-0D (left), and another to predict the F1-optimal threshold~\cite{f1score} (right). Results on artificial data are shown in green and on real-world data in purple.}
    \label{fig:failures}
\end{figure}

Beyond severity and uncertainty, FoMo-X can, in principle, support many
additional diagnostic readouts trained on frozen embeddings. However, our
experiments show that validating their performance is a crucial step, as generalization is not guaranteed.

We trained an auxiliary head to estimate the outlier-detection performance of
the frozen FoMo-0D model on a dataset, using AUROC~\cite{rocauc} as the
target. Such a head would be attractive for deployment because it could serve as
a dataset-level reliability indicator and could enable automated model selection
and pipeline decisions (e.g., selecting preprocessing steps) without ground-truth
labels. In our implementation, the head produces a per-sample prediction that is
aggregated (averaged) into a single dataset-level estimate. As shown in
Figure~\ref{fig:rocauc}, this approach yields strong correlation on synthetic test tasks
($\rho_s \approx 0.95$), but the correlation collapses on ADBench
($\rho_s \approx -0.01$), indicating a near-complete failure to transfer.

This failure suggests that the head exploits regularities that are specific to the
simulator prior and do not hold for real datasets.

From our experiments, it seems that heads predicting \emph{dataset-level} (global) properties
are substantially harder to train than heads predicting \emph{per-sample}
(local) diagnostics.

This can also be seen in a related experiment that targeted a practically important, but similarly global,
quantity: the decision threshold that maximizes the F1-score~\cite{f1score}.
While AUROC/AUPRC~\cite{aucpr,rocauc} summarize ranking quality, operational OD deployments
often require a binary decision (``outlier'' vs.\ ``normal''), making threshold
selection a central bottleneck in unsupervised settings. On synthetic tasks, the
threshold head achieves a high correlation with the oracle (label-derived) optimum
($\rho_s \approx 0.95$), but transfer again degrades on ADBench
($\rho_s \approx 0.60$). Interestingly, we observe a systematic bias in absolute threshold
values (Figure~\ref{fig:thresh}), with predicted thresholds that are usually too small.

Such systematic deviation is consistent with a calibration mismatch
between the simulator prior and real-world datasets: even if the head preserves
some relative ordering of ``easier'' vs.\ ``harder'' threshold regimes, the absolute
operating point shifts. While Figure~\ref{fig:thresh} illustrates that a simple linear
transformation can partially correct the bias, such a correction requires at least some labeled
real-world data and should therefore be viewed as a post-hoc calibration step. In a
realistic evaluation protocol, any calibration must be learned on held-out
datasets (or a separate calibration split) to avoid test-set leakage. Even with an unrealistically strong correction and possible test-set leakage, we still observe a significant drop in performance.

\begin{figure}[t]
    \centering
    \includegraphics[width=0.5\linewidth]{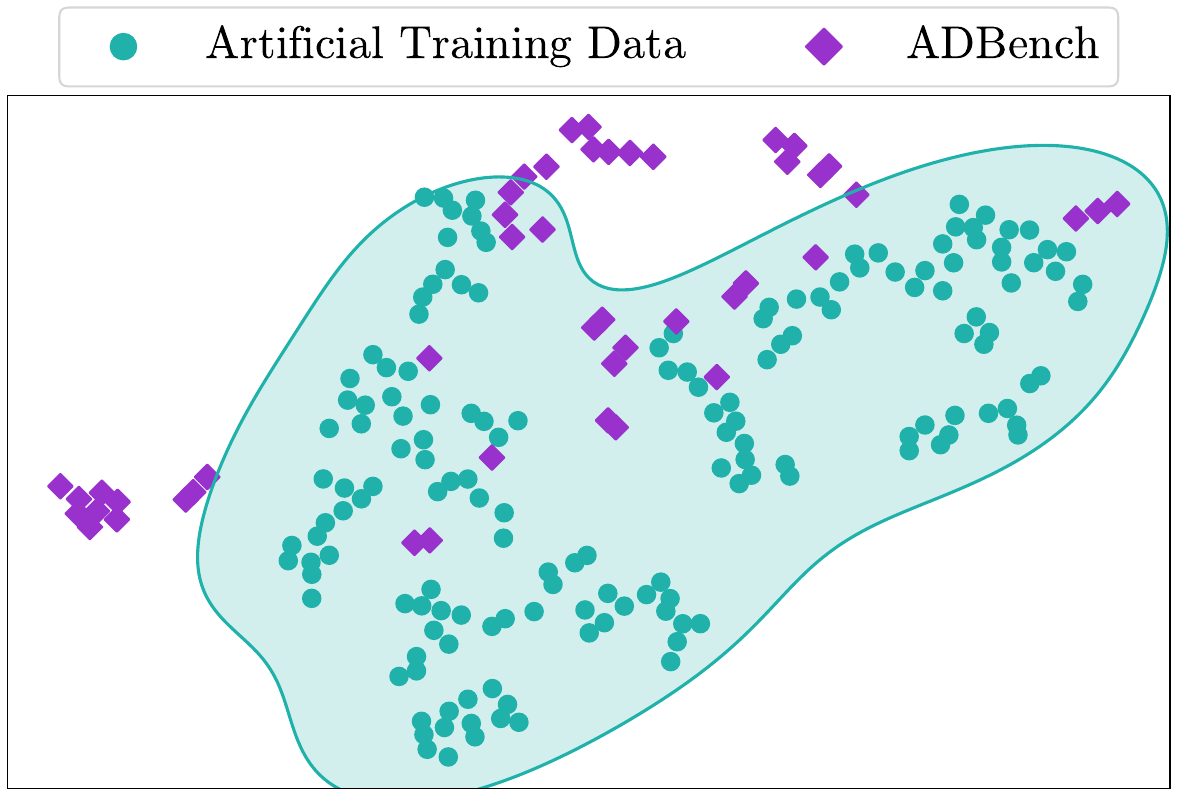}
    \caption{t-SNE embedding of ADBench (purple) and artificial datasets (cyan) used for foundation model training based on dataset-level meta-features \cite{Vanschoren2018MetaLearningAS}. We visualize the distribution of the training datasets using kernel density estimation~\cite{kde}, with the density region chosen to cover all training datasets. \textbf{A large number of real-world datasets are well separated from the training data used to train our foundation model heads.}}
    \label{fig:overlap}
\end{figure}

\subsection{Modular Outlier Foundation Models}

The transfer gaps observed for global heads point to broader limitations of
current outlier foundation models and motivate directions for improvement.

FoMo-0D and our heads are trained under a simulator-induced prior based on
GMM-style generative mechanisms. While this prior enables large-scale training
without labeled real outliers, it is necessarily a simplification of real-world
tabular OD. The fact that we observe a systematic bias in the threshold head's predictions suggests a difference between artificial and real-world datasets.  Figure~\ref{fig:overlap} supports this mismatch qualitatively: a t-SNE embedding of
dataset-level meta-features~\cite{Vanschoren2018MetaLearningAS} shows that many ADBench
datasets occupy regions that are weakly covered by the synthetic training
distribution. This separation is consistent with the systematic deviations we
observe when transferring certain heads, and suggests that the simulator prior
does not span important real-world dataset characteristics (e.g., complex
dependencies, heavy tails, heterogeneous feature scales, and non-Gaussian
structures; see Figure~\ref{fig:one}).

A direct consequence is that improving the \emph{data prior} is central for scaling
FoMo-X beyond the heads studied here. Promising options include richer,
more expressive priors~\cite{fomo2Outformer,tabPFN2dot5} and continued pretraining or
adaptation using real-world tabular data~\cite{tabfnRW}, analogous to recent
advances in supervised tabular foundation models. For outlier detection, the
current limitation is the relatively small number of curated real OD datasets for
such adaptation (e.g., the 47 native-tabular ADBench datasets). Nevertheless,
our training curves indicate that some heads saturate after few epochs
(Figure~\ref{fig:progress}), which suggests that head-level adaptation on real data may be feasible
even with moderate dataset collections, provided that the evaluation is performed on
previously unseen datasets.

A second limitation concerns the architecture of current outlier foundation
models. FoMo-0D embeds all raw features through a single dense projection into
a fixed token space, which enforces a hard feature limit (100 features) and 
mixes feature identity early, making it difficult to attach auxiliary heads that
produce faithful, column-aligned feature-level explanations (e.g., feature importances~\cite{FI4AD}, shapley values~\cite{deanshap}, or counterfactuals~\cite{counterfactual4AD}). While post-hoc
feature attribution methods remain applicable in principle, a more direct
integration of feature-centric explainability would likely require architectures
that preserve column identity explicitly (e.g., column-wise tokenization or
structured per-feature embeddings) and scale beyond the current truncation
regime. Existing work mitigates the 100-feature limit via ensembling across
feature subsets~\cite{fomo2Outformer,feature-bagging}, but this increases runtime,
complicates explanation methods and further mixes feature identities, thus also preventing feature-level explanations.

Besides explainability, the modular setup of FoMo-X can also be used to add further heads for diverse use cases. This is particularly relevant because OD covers heterogeneous problem settings and operational goals~\cite{positionpaperRoechnerBenchmarking}, which are not always well served by a single out-of-the-box detector~\cite{activepersonalisation}. Thus, outlier predictions specialized for outliers, anomalies, or novelties~\cite{typesOfAnomalies}, for global or local deviations~\cite{boumanHowMany}, or with different requirements like fairness~\cite{lemanFairness} now only require artificial training datasets to add specialized heads, provided that suitable simulator targets and robust real-world evaluation protocols can be defined. 

Overall, FoMo-X suggests a practical direction for trustworthy OD foundation
models: expose a selection of single-pass diagnostic heads that decode
uncertainty- and triage-relevant signals from frozen representations. Our results
also highlight that head design must explicitly consider transfer under prior
shift. Heads that estimate global dataset properties appear particularly fragile,
whereas local diagnostics aligned with the backbone’s inference mechanism
transfer more reliably. Closing the simulator--real gap, and adopting
feature-structured architectures, are therefore key steps toward broader and
more faithful explainability in modular outlier foundation models.

\section{Conclusion}
In this paper, we introduce \textbf{FoMo-X}, a modular framework that augments foundation models for outlier detection with lightweight, self-explanatory diagnostic signals. Building on FoMo-0D~\cite{fomo}, FoMo-X attaches auxiliary heads to frozen PFN embeddings, enabling the model to explain its own predictions without altering detection behavior, while requiring negligible computational cost.

We instantiated FoMo-X with two heads that capture epistemic uncertainty and outlier severity, and are trained solely on simulator-derived supervision. Experiments show that both signals generalize meaningfully to real-world benchmarks: the uncertainty head accurately approximates expensive dropout-based variability in a single pass, while the severity head provides a reliable tiering that correlates with prediction errors and decision boundary ambiguity.

Our analysis also reveals limitations. Not all explainability targets transfer equally well from synthetic priors to real data—heads predicting global, dataset-level properties exhibit strong prior dependence and limited generalization.

Overall, FoMo-X demonstrates that foundation models for outlier detection can be trained to explain themselves efficiently, without affecting inherent predictions. We view this as a first step towards richer, modular outlier foundation models.

\begin{credits}
\subsubsection{\ackname}
This research was in part funded by the Research Center Trustworthy Data Science and Security (\url{https://rc-trust.ai}), one of the Research Alliance centres within the University Alliance Ruhr (\url{https://uaruhr.de}).

The research was further supported with computing time provided on LiDO3 (\url{https://lido.itmc.tu-dortmund.de/lido3/}).

\subsubsection{Declaration on Generative AI.}
During the preparation of this article, we used the ChatGPT 5.2 model from OpenAI to preselect additional related work and for language edits, aiming to enhance readability.
After using this service, the authors reviewed and edited the content as needed and take full responsibility for the manuscript's content.

\subsubsection{\discintname}
The authors have no competing interests to declare that are relevant to the content of this article.

\end{credits}

%
%
%
\bibliographystyle{splncs04}
\bibliography{fomox}

\end{document}